\documentclass{article}
\usepackage{ijcai20}
\usepackage{xcolor}
\usepackage{comment}
\usepackage{times}
\usepackage{soul}
\usepackage{bbold}
\usepackage{url}
\usepackage[hidelinks]{hyperref}
\usepackage[utf8]{inputenc}
\usepackage[small]{caption}
\usepackage{graphicx}
\usepackage{amsmath}
\usepackage{amsthm}
\usepackage{booktabs}
\usepackage{algorithm}
\usepackage{amsfonts}
\usepackage{amssymb}
\usepackage[noend]{algpseudocode}
\usepackage{changepage}

\title{Disentangled Variational Autoencoder based Multi-Label Classification with Covariance-Aware Multivariate Probit Model}

\author{
Junwen Bai\footnote{Contact Author}
\and
Shufeng Kong\And
Carla Gomes
\affiliations
Department of Computer Science, Cornell University
\emails
\{jb2467, sk2299\}@cornell.edu,
gomes@cs.cornell.edu
}

\begin{document}

\maketitle

\begin{abstract}
  Multi-label classification is the challenging task of predicting the presence and absence of multiple targets, involving representation learning and label correlation modeling. We propose a novel framework for multi-label classification, Multivariate Probit Variational AutoEncoder (MPVAE), that  effectively learns latent embedding spaces as well as label correlations. MPVAE learns and aligns two probabilistic embedding spaces for labels and features respectively. The decoder of MPVAE takes in the samples from the embedding spaces and models the joint distribution of output targets under a Multivariate Probit model by learning a shared covariance matrix. We show that MPVAE outperforms the existing state-of-the-art methods on 
  a variety of
  application domains, using public real-world datasets\footnote{Our code is available on https://github.com/JunwenBai/MPVAE}. MPVAE is further shown to remain robust under noisy settings. Lastly, we demonstrate the interpretability of the learned covariance by a case study on a bird observation dataset.
\end{abstract}

\section{Introduction}

Multi-label classification (MLC) {concerns the simultaneous prediction of the presence and absence of multiple  labels for each sample of a given sample set.}
Unlike in the conventional classification task, more than one label or target could be associated with each sample in MLC \cite{zhang2013review,zhang2018binary}. This setting is important for the study of 
a variety of scenarios, such as joint species distributions mapping \cite{chen2016deep},
protein site localization \cite{alazaidah2015multi} and drug side effects \cite{kuhn2015sider}. Furthermore, understanding the label correlations is also important. For instance, in biodiversity applications, species correlation modeling is critical to address core ecological concerns like interactions of species with each other, which could affect species monitoring, protection and policy-making \cite{evans2017species}.

Some early work simply decomposes the multi-label classification into multiple single-label classification problems \cite{boutell2004learning}. Though these methods can be adapted from single-label predictors, they ignore the correlation among labels. To improve this, classifier chains \cite{read2009classifier} stack the binary classifiers into a chain and reuse the outputs of previous classifiers as extra information to improve the prediction of the current label. Followup works extend the classifier chains to recurrent neural networks \cite{wang2016cnn} to increase capacity and better model the label correlation. Label ordering is critical to these methods since long-term dependencies are typically weaker than short-term dependencies. The model structure also restricts parallel computation. Another straightforward method is to find nearest neighbors in the feature space and assign labels to test samples by Bayesian inference \cite{zhang2007ml,chiang2012ranking}. However, either the predefined metric space or the prior may heavily affect the model performance.

Latent embedding learning is a recent technique to match features and labels in the latent space. Pioneer studies \cite{yu2014large,chen2012feature,bhatia2015sparse} make low-rank assumptions of labels and features, and transform labels to label embeddings, by dimensionality reduction techniques such as canonical correlation analysis (CCA). Benefiting from the capacity of deep neural networks, more recent latent embedding methods for MLC employ neural networks to build and align the latent spaces for both labels and features \cite{yeh2017learning,chen132019two}. The constraints in the conventional dimension reduction models are relaxed and embedded into the deep latent space. For example, C2AE relaxes the orthogonality constraint to the minimization of an $\ell_2$ distance. These embedding methods are believed to implicitly encode the label correlations in the embedding space. 

Some other state-of-the-art models initiate the research on using graph neural networks (GNN) to explicitly encode the label correlations \cite{chen2019multi,lanchantin2019neural}. A graph neural network for labels can build dependencies among labels through learned or given edges between them. Though GNN brings a new way to embed the correlations, the number of stacked GNNs or the iterations of message passing may require extra effort to fine-tune.

{
{We propose the
\textbf{Multivariate Probit Variational Autoencoder (MPVAE)}, which \textit{improves both the embedding space learning and label correlation encoding}.
In particular,  \textbf{(1)} MPVAE learns probabilistic latent spaces for both labels and features, unlike most autoencoder (AE) based multi-label models.} The probabilistic latent space learned by the VAE can provide three major advantages. First, it gives more control to the latent space \cite{chung2015recurrent}. In many AE models, one can often observe the label-encoder-decoder branch gives much better performance than feature-encoder-decoder branch. Imposing the VAE structure in the latent space {helps  balance} the difficulty of the learning and aligning of the two subspaces. Second, smoothness in the latent space is often desired   \cite{wu2018multi}. Probabilistic models like the VAE naturally bring smoothness on a local scale since the decoder decodes a sample rather than a specific embedding. Third, the VAE model and its variations learn representations with \textbf{disentangled} factors \cite{van2019disentangled}. If both latent spaces for features and labels learn disentangled factors, not only is it  helpful to aligning two spaces, but it is also beneficial for the decoding process. \textbf{(2) MPVAE explicitly learns a shared \iffalse low-rank \fi covariance matrix to build dependencies among labels} by adopting the Multivariate Probit (MP) probabilistic model, which is  inspired by some recent work in joint distribution modeling \cite{chen2018end}. 
The MP assumes an underlying latent multivariate Gaussian distribution.
We show that the MP model is a simple and straightforward component of the overall probabilistic generative framework compared to other more complex models such as GNNs. More importantly, the MP model improves the prediction performance and provides the interpretability of the learned covariance matrix. By using the Cholesky decomposition and t-SNE, we demonstrate visually the value of the learned covariance, in applications like species correlation modeling.
\textbf{(3) MPVAE is optimized with respect to  a three-component loss function}, which includes a Kullback–Leibler (KL) divergence component for jointly learning and aligning the label and feature embeddings and a cross-entropy and ranking loss for the multi-label prediction in the MP model. \textbf{(4)} We thoroughly test MPVAE with multiple public datasets on a variety of metrics. \textbf{MPVAE outperforms 
(or is comparable to)
other state-of-the-art multi-label prediction models.}
  We further illustrate MPVAE is still robust even if the training labels are noisy. 
}

\section{Other Related Work}

Covariance matrices are commonly seen in non-deep multi-label classification models \cite{bi2014multilabel,zhang2013multilabel}. Non-deep models often assume a matrix-variate normal distribution on features, weights or labels. Covariances thus play a key role in these models. But the scalability issue limits their applications in large-scale problems. However, these assumptions are not necessary in deep neural networks given their powerful expressiveness.

A recent success of combining deep learning and covariance based paradigms is the deep Multivariate Probit model (DMVP) \cite{chen2018end}. It is a deep generalization of the classic Multivariate Probit model. An efficient sampling process was proposed in the paper to avoid the heavy-duty Markov chain Monte Carlo (MCMC) sampling process. 
Though DMVP performs well on the joint likelihood measure, it lacks enough predictive power for presence-absence (0/1) classification.
MPVAE makes one step further and introduces the cross-entropy loss as well as the ranking-loss under the Multivariate Probit paradigm. 

Some prior work also applied deep generative models in multi-label classification. \cite{chu2018deep} proposes a deep sequential generative model. Though the model is effective in the setting of missing labels, the concern of unstable training of stacked generative models should not be overlooked.

\section{Methods}

\begin{figure*}[t]
    \centering
    \includegraphics[width=0.9\textwidth]{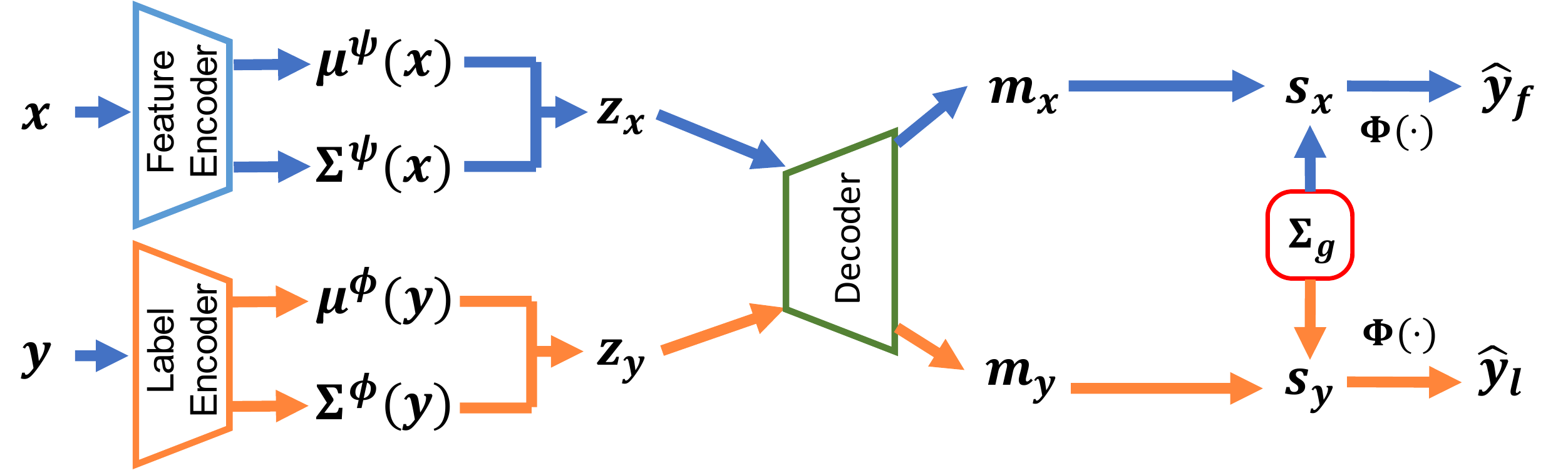}
    \caption{Network structure of MPVAE. The feature encoder encodes $x$ to a probabilistic latent subspace with a neural network parameterized by $\psi$. Similarly, another label encoder with parameter $\phi$ maps $y$ to another probabilistic latent subspace with the same dimensionality. Two samples $z_x,z_y$ from the subspaces are fed into the shared decoder and deciphered as the means $m_x$, $m_y$ in the Multivariate Probit model. With the help of the global covariance matrix $\Sigma_g$, we sample $s_x$, $s_y$ from $\mathcal N(m_x,\Sigma_g),\mathcal N(m_y,\Sigma_g)$ to derive the final 0/1 predictions $\hat{y}_f,\hat{y}_l$. Note that during testing, only $\hat{y}_f$ is the prediction for the test instances.}
    \label{fig:mpvae}
\end{figure*}

\subsection{Preliminaries}

Let $\mathcal D$ denote the dataset $\{(\textbf{x}_i,\textbf{y}_i)\}_{i=1}^N$, where $\textbf{x}_i\in \mathbb R^S$ and $\textbf{y}_i\in \{0,1\}^L$. $\textbf{x}_i$ is the feature vector and $\textbf{y}_i$ represents the presence (1) or absence (0) of targets.

\subsubsection{Variational Autoencoder (VAE)}
A VAE assumes a generative process for the observed datapoints $X$: $P(X)=\int p_\theta(X|z;\theta)P(z)dz$, by introducing latent variables $z$. Since most $z$'s contribute little to $P(X)$, Monte Carlo sampling would be inefficient. We instead learn a function $q_\phi(z|X)$ to approximate the intractable $P(z|X)$ for efficient sampling. The KL divergence ($\mathcal D$) between $q_\phi(z|X)$ and $P(z|X)$ is given by
$\mathcal D(q_\phi(z|X)||P(z|X))=\mathbb E_{z\sim q_\phi}[\log q_\phi(z|X)-\log P(z|X)]$. Applying Bayes rule and transforming the equation yield: 
$\log P(X)-\mathcal D[Q(z|X)||P(z|X)]=\mathbb E_{z\sim q_\phi}[\log p_\theta(X|z)]-\mathcal D[q_\phi(z|X)||P(z)]$.
The right hand side of the equation is the tractable evidence lower bound (ELBO) to maximize. $P(z)$ is the prior, a standard multivariate normal distribution. The first term in the ELBO encourages the reconstruction of $X$ and the second term penalizes the KL divergence between the approximate distribution and the prior to impose structure on the latent space. Both $p_\theta$ and $q_\phi$ can be parameterized by neural networks. With the help of {the} reparameterization trick, the whole model can be trained with back-propagation. VAE and its variations can learn disentangled factors by controlling the capacity of {the} information bottleneck. For example, $\beta$-VAE \cite{higgins2017beta} is a known disentangled VAE, which is able to learn abstract concepts like size and shape, only with a slight modification to the objective,
$\mathbb E_{z\sim q_\phi}[\log p_\theta(X|z)]-\beta\mathcal D[q_\phi(z|X)||P(z)]$. MPVAE follows the structure of $\beta$-VAE with $\beta=1.1$.

\subsubsection{Multivariate Probit (MP) Model}
Consider a single sample $(\textbf{x}, \textbf{y})$ where $\textbf{x}$ is the input feature vector and $\textbf{y} \in \{0,1\}^L$ is the label. The Multivariate Probit model introduces auxiliary latent variables $\textbf{y}^* \in \mathbb R^L$, which follow a multivariate normal distribution $\mathcal N(\textbf{x}\gamma, \Sigma)$ where $\gamma$ is the weight parameter and $\Sigma$ is the covariance matrix. $\textbf{y}$ is viewed as the indicator for whether $\textbf{y}^*$ is positive or not: $y_i=\mathbb 1\{y^*_i>0\},i=1,..,L$. The probability of observing $\textbf{y}$ is given by $P(\textbf{y}|\textbf{x}\gamma,\Sigma)=\int_{A_L}...\int_{A_1}p(\textbf{y}^*|\textbf{x}\gamma,\Sigma)dy^*_1...dy^*_L$, where $A_j=(-\infty,0]$ if $y_j=0$ or $A_j=(0,\infty)$ otherwise. $p(\cdot)$ is the probability density function of the normal distribution. The framework can be generalized to a deep model simply by replacing the mean $\textbf{x}\gamma$ with a neural network $f(\textbf{x})$. In MPVAE, the mean is given by the decoder of VAE. %

\subsection{MPVAE}

We propose Multivariate Probit Variational Autoencoder, a novel disentangled Variational Autoencoder based framework with covariance-aware Multivariate Probit model for MLC. The illustration 
of the framework is shown in Fig. \ref{fig:mpvae}. The whole model can be viewed as a two-stage generative process. The first stage maps the features and labels to Gaussian subspaces where the means and variances are learned by multi-layer perceptrons. The key task in this stage is to match the two subspaces. The second stage decodes the sample from each subspace and feeds the outputs into a Multivariate Probit module as the means. A global covariance matrix is learned separately. The final output of the second stage gives the predicted labels.

\subsubsection{Learning and Aligning Probabilistic Subspaces}

Given an input pair of feature vector and label $(\textbf{x},\textbf{y})$ where $\textbf{x}\in \mathbb R^S, \textbf{y} \in \{0,1\}^L$, the  feature encoder maps $\textbf{x}$ to a Gaussian subspace $\mathcal N(\mu^\psi(\textbf{x}), \Sigma^\psi(\textbf{x}))$ and the label encoder maps $\textbf{y}$ to another Gaussian subspace $\mathcal N(\mu^\phi(\textbf{y}), \Sigma^\phi(\textbf{y}))$. $\phi,\psi$ are trainable parameters in the encoders.  $\mu^\psi(\textbf{x}),\mu^\phi(\textbf{y})\in \mathbb R^d$ and $\Sigma^\psi(\textbf{x}),\Sigma^\phi(\textbf{y})\in \mathbb R_{\ge 0}^{d\times d}$, where $d$ is the dimensionality of the latent space. $\textbf{z}_\textbf{x}, \textbf{z}_\textbf{y}$ denote samples from each of these two distributions respectively.

If we only consider the label encoder-decoder branch (orange branch in Fig. \ref{fig:mpvae}), it models a standard ($\beta$-)VAE. The ELBO to optimize can be written as,
\begin{equation}
    \mathbb E_{\textbf{z}\sim q_\phi}[\log p_\theta(\textbf{y}|\textbf{z})]-\beta\mathcal D[q_\phi(\textbf{z}|\textbf{y})||P(\textbf{z})]
\end{equation}
The issue with this label autoencoder is the lack of connections between $\textbf{x}$ and $\textbf{z}$. Even if a good generative model is learned, prediction given $\textbf{x}$ is impossible since the prior $P(\textbf{z})$ is unrelated to $\textbf{x}$. Our simple fix is to replace $P(\textbf{z})$ with a prior distribution dependent on $\textbf{x}$, $q_\psi(\textbf{z}|\textbf{x})$. That's where the feature encoder comes from. The feature encoder is also a neural network parameterized by learnable $\psi$. With the feature encoder, given the input $\textbf{x}$, we can sample from $q_\psi(\textbf{z}|\textbf{x})$, which is approximately equal to $q_\theta(\textbf{z}|\textbf{y})$. The challenges thereafter are twofold: a) how to learn $\psi$ and b) how to align $\mathcal N(\mu^\psi(\textbf{x}), \Sigma^\psi(\textbf{x}))$ and $\mathcal N(\mu^\phi(\textbf{y}), \Sigma^\phi(\textbf{y}))$. We propose to extend the objective function to:
\begin{small}
\begin{equation*}
    \frac{1}{2}(\mathbb E_{\textbf{z}\sim q_\phi}[\log p_\theta(\textbf{y}|\textbf{z})]+\mathbb E_{\textbf{z}\sim q_\psi}[\log p_\theta(\textbf{y}|\textbf{z})])-\beta\mathcal D[q_\phi(\textbf{z}|\textbf{y})||q_\psi(\textbf{z}|\textbf{x})]
\end{equation*}
\end{small}
The first two terms will be handled by the Multivariate Probit model in the next subsection. The last term is simply the KL divergence between two multivariate normal distributions. 
Since both distributions have diagonal covariance matrices, we can derive the KL loss term for $\mathcal D[q_\phi(\textbf{z}|\textbf{y})||q_\psi(\textbf{z}|\textbf{x})]$:
\begin{small}
\begin{equation}\label{kl}
\begin{split}
    L_{\text{KL}}(\textbf{x},\textbf{y})=&\frac{1}{2}\beta[\sum_{i=1}^d\log\frac{\Sigma^\psi_{i,i}(\textbf{x})}{\Sigma^\phi_{i,i}(\textbf{y})}-d+\sum_{i=1}^d\frac{\Sigma^\phi_{i,i}(\textbf{y})}{\Sigma^\psi_{i,i}(\textbf{x})}+\\
    &\sum_{i=1}^d\frac{(\mu^\psi_i(\textbf{x})-\mu^\phi_i(\textbf{y}))^2}{\Sigma^\psi_{i,i}(\textbf{x})}]
\end{split}
\end{equation}
\end{small}
As in the conditional VAE, our implementation further adds an improvement to concatenate each of $\textbf{y}, \textbf{z}_\textbf{y}, \textbf{z}_\textbf{x}$ with $\textbf{x}$, which basically does not affect the derivations above, but uses extra information for inference.

\subsubsection{Reconstruction and Prediction}

The Multivariate Probit (MP) is a classic latent variable model for data with presence-absence relationships. Unlike the typical softmax transformation in most deep methods, the model maps a sample from a multivariate normal distribution to its own cumulative distribution (CDF), to bound the output range within $[0,1]$. 
The major drawback of the MP model is that the integration step is intractable for large-scale data and is usually approximated with MCMC.
\cite{chen2018end} provides an alternative parallelizable sampling process for estimating the CDF. GPUs can thus expedite the estimation. We adopt this sampling method in our model, and find it works well in MPVAE.

Label information is available for both the training and testing phases in \cite{chen2018end}. However, as a predictive model, MPVAE does not have access to labels until the final predicted targets are given, when the loss can be computed between the ground-truth and predicted labels for training, or the predicted targets are directly given as the output during testing. In this case, we let MPVAE calculate the integral only w.r.t. the $(0, \infty)$ region for each target(dimension). If the corresponding target is present, the CDF should be close to 1. Otherwise, it should be near 0.

\paragraph{Sampling Process.} The shared decoder reads the samples $\textbf{z}_\textbf{x}$,  $\textbf{z}_\textbf{y}$, and outputs $\textbf{m}_\textbf{x}$, $\textbf{m}_\textbf{y}\in \mathbb R^L$ respectively as the means in the MP model. The covariance matrix $\Sigma_g\in \mathbb R^{L\times L}$ in the MP only models the dependencies among targets and is unrelated to the features. Thus $\Sigma_g$ can be learned as a shared parameter. This 
stabilizes the training and helps interpret the label correlation (shown in experiments).
Instead of directly sampling $\textbf{y}^*$ from $\mathcal N(\textbf{m},\Sigma_g)$, we sample twice by decomposing $\Sigma_g$ to $V+\Sigma_r$ where $V$ is a diagonal positive definite matrix and $\Sigma_r$ is the residual. For 2 random variables $\textbf{w}\sim\mathcal N(\textbf{0},V),\textbf{s}\sim\mathcal N(\textbf{m},\Sigma_r)$, the difference  $(\textbf{w}-\textbf{s})\sim\mathcal N(-\textbf{m},\Sigma_g)$. Note that since $V$ is diagonal, the different dimensions of $\textbf{w}$ are independent. To estimate the probability of presence $P(y^*_i\ge 0|\textbf{m},\Sigma_g), i\in[1,L]$,
\begin{equation}
\begin{split}
    &P(y^*_i\ge 0|\textbf{m},\Sigma_g)=P(y^*_i\le 0|-\textbf{m},\Sigma_g)\\
    =&P(w_i-s_i\le 0)~~~ \textbf{w}\sim\mathcal N(\textbf{0},V),~\textbf{s}\sim\mathcal N(\textbf{m},\Sigma_r)\\
    =&\mathbb E_{\textbf{s}\sim\mathcal N(\textbf{m},\Sigma_r)}[P(w_i\le s_i|\textbf{s})]~~~\textbf{w}\sim\mathcal N(\textbf{0},V)\\
    =&\mathbb E_{\textbf{s}\sim\mathcal N(\textbf{m},\Sigma_r)}[\Phi(\frac{s_i}{\sqrt{V_{i,i}}})]
\end{split}
\end{equation}
where $\Phi$ is the CDF of a univariate standard normal distribution. Since $V$ is simply a scaling factor, w.l.o.g., $V$ can be set to identity $I$. Under this mechanism, $P(y^*_i\ge 0|\textbf{m},\Sigma_g)$ for each $i$ can be computed in parallel and independently with one sample $\textbf{s}$ or the average of multiple samples. Note that with this derivation, $\Sigma_r$ is learned rather than $\Sigma_g$. But they only differ by $I$. The 0/1 outputs given $\textbf{m}_\textbf{x},\textbf{m}_\textbf{y}$ are denoted by $\hat{\textbf{y}}_f$ and $\hat{\textbf{y}}_l$. If $P(y^*_i\ge 0|\textbf{m},\Sigma_g)$ is higher than a certain threshold, $\hat{y}_i$ gives 1. Otherwise, $\hat{y}_i$ is set to 0. During testing, $\hat{\textbf{y}}_f$ is regarded as the final result. 

\paragraph{Binary Cross Entropy (BCE) Loss in MP.} 
With the efficient sampling scheme, the reconstruction losses $\mathbb E_{\textbf{z}\sim q_\phi}[\log p_\theta(\textbf{y}|\textbf{z})]$ and $\mathbb E_{\textbf{z}\sim q_\psi}[\log p_\theta(\textbf{y}|\textbf{z})]$ can be concretized. For a binary prediction task for each target, a Bernoulli likelihood assumption is valid, which leads to the binary cross entropy loss between the labels and predictions:
\begin{small}
\begin{equation*}
\begin{split}
    &\mathcal L_{\text{CE}}(\textbf{y},\textbf{z})=-\log p_\theta(\textbf{y}|\textbf{z},\Sigma_g)=-\log p_\theta(\textbf{y}|\textbf{m},\Sigma_r)\\
    =&-\log \mathbb E_{\textbf{s}\sim\mathcal N(\textbf{m},\Sigma_r)}[\prod_{i=1}^L\Phi(s_i)^{y_i}(1-\Phi(s_i))^{1-y_i}]\\
    =&-\log \mathbb E_{\textbf{s}\sim\mathcal N(\textbf{m},\Sigma_r)}\exp[{\sum_{i=1}^L}y_i\log\Phi(s_i)+(1-y_i)\log(1-\Phi(s_i))]\\
    \approx&-\log \frac{1}{M}\sum_{k=1}^M\exp[{\sum_{i=1}^L}y_i\log\Phi(s_i^k)+(1-y_i)\log(1-\Phi(s_i^k))]
\end{split}
\end{equation*}
\end{small}
The last step is a Monte Carlo approximation. $M$ is the preset number of samples.
The log-sum-exp trick is applied for computation to avoid overflow issues. 
Since $\mathbb E_{\textbf{z}\sim q_\phi}[\log p_\theta(\textbf{y}|\textbf{z})]$ is typically approximated by a single sample from $q_\phi$ and minimization is preferred for training, the objective function can be written as
$-\mathbb E_{\textbf{z}\sim q_\phi}[\log p_\theta(\textbf{y}|\textbf{z})]\approx -\log p_\theta(\textbf{y}|\textbf{z}_\textbf{y})=\mathcal L_{\text{CE}}(\textbf{y},\textbf{z}_\textbf{y})$ where $\textbf{z}_\textbf{y}\sim q_\phi(\textbf{z}|\textbf{y})$. Similarly, $-\mathbb E_{\textbf{z}\sim q_\psi}[\log p_\theta(\textbf{y}|\textbf{z})]\approx -\log p_\theta(\textbf{y}|\textbf{z}_\textbf{x})=\mathcal L_{\text{CE}}(\textbf{y},\textbf{z}_\textbf{x})$ where $\textbf{z}_\textbf{x}\sim q_\psi(\textbf{z}|\textbf{x})$.

\paragraph{Ranking Loss in MP.} Ranking loss \cite{zhang2013review} is also widely used in many multi-label tasks. It is a loss to measure the correlations between positive labels and negative labels. The idea behind the ranking loss is simple: the gap between the logits for positive and negative labels should be as large as possible. Suppose $\textbf{y}$ is the ground-truth label set. Let $\textbf{y}^0$ denote the set of indices of negative labels and $\textbf{y}^1$ the positive labels. $\textbf{s}$ denotes the sample from $\mathcal N(\textbf{m},\Sigma_r)$, which depends on $\textbf{z}$. Thus the ranking loss on $\textbf{s}$ can also be viewed as a loss on $\textbf{z}$. We define the ranking loss in MP as
\begin{small}
\begin{equation*}
\begin{split}
    &\mathcal L_{\text{RL}}(\textbf{y},\textbf{z})=\mathcal L_{\text{RL}}'(\textbf{y},\textbf{s})\\
    =&\mathbb E_{\textbf{s}\sim\mathcal N(\textbf{m},\Sigma_r)}[\frac{1}{|\textbf{y}^0||\textbf{y}^1|}\sum_{(i,j)\in\textbf{y}^1\times\textbf{y}^0}\exp(-(\Phi(s_i)-\Phi(s_j)))]\\
    =&\frac{1}{M}\sum_{k=1}^M[\frac{1}{|\textbf{y}^0||\textbf{y}^1|}\sum_{(i,j)\in\textbf{y}^1\times\textbf{y}^0}\exp(-(\Phi(s_i^k)-\Phi(s_j^k)))]
\end{split}
\end{equation*}
\end{small}
The ranking loss can be defined for both $\textbf{z}_\textbf{y}$ and $\textbf{z}_\textbf{x}$.

\paragraph{Entropy Loss in MP.} Some multi-label datasets have very sparse positive labels even if there exist multiple labels for one feature. For example, in the \textit{mirflickr} dataset, the positive label rate is as low as $12\%$. Therefore, for sparse datasets, we add an extra entropy loss (define $p_i=\frac{\exp(\Phi(s_i))}{\sum_{j=1}^L\exp(\Phi(s_i))}$), 

\begin{small}
\begin{equation}
\mathcal L_{\text{Ent}}(\textbf{z})=\mathcal L_{\text{Ent}}'(\textbf{s})=\mathbb E_{\textbf{s}\sim\mathcal N(\textbf{m},\Sigma_r)}[-\sum_{i=1}^L p_i\log p_i]
\end{equation}
\end{small}
Entropy loss acts as a self-regularizer and only depends on the predicted values. 

\subsubsection{Overall Loss Function} 

\begin{algorithm}[tbp]
\caption{Training MPVAE}
\label{alg:algorithm}
\textbf{Input}: $\{(\textbf{x}_i,\textbf{y}_i)\}_{i=1}^N$, batch size $b$
\begin{adjustwidth}{-0.55em}{}
\begin{algorithmic}[1] 
\For{\# of iterations}
    \State Sample a minibatch $\{(\textbf{x}^{(k)},\textbf{y}^{(k)})\}_{k=1}^b$
    \State Compute $\{(\mu^\psi(\textbf{x}^{(k)}),\Sigma^\psi(\textbf{x}^{(k)})),(\mu^\phi(\textbf{y}^{(k)}),\Sigma^\phi(\textbf{y}^{(k)}))\}$
    \State Derive $\mathcal L_{\text{KL}}$ by Eq (\ref{kl})
    \State Sample latent variables $\{\textbf{z}_\textbf{x}^{(k)}\},\{\textbf{z}_\textbf{y}^{(k)}\}$
    \State Decode $\textbf{z}$'s and obtain $\{\textbf{m}_\textbf{x}^{(k)}\},\{\textbf{m}_\textbf{y}^{(k)}\}$
    \State Calculate the mean loss by Eq (\ref{total}) for the batch
    \State Compute gradients and update parameters with Adam
\EndFor\\
\Return $\phi,\psi,\theta,\Sigma_r$
\end{algorithmic}
\end{adjustwidth}
\end{algorithm}

Let $\mathcal L_{prior}=\lambda_1\mathcal L_{\text{CE}}(\textbf{y},\textbf{z}_\textbf{x})+\lambda_2 L_{\text{RL}}(\textbf{y},\textbf{z}_\textbf{x})+\lambda_3 L_{\text{Ent}}(\textbf{z}_\textbf{x})$ and $\mathcal L_{recon}=\lambda_1\mathcal L_{\text{CE}}(\textbf{y},\textbf{z}_\textbf{y})+\lambda_2 L_{\text{RL}}(\textbf{y},\textbf{z}_\textbf{y})+\lambda_3 L_{\text{Ent}}(\textbf{z}_\textbf{y})$. Together with the KL loss $\mathcal L_{\text{KL}}$ defined in the previous subsection, the overall loss function can be described as
\begin{equation}\label{total}
\mathcal L=\mathcal L_{\text{recon}}+\mathcal L_{\text{prior}}+\beta\mathcal L_{\text{KL}}
\end{equation}
$\mathcal L_{\text{prior}}$ encompasses the losses in the feature-encoder-decoder branch (blue branch) for the prior, while $\mathcal L_{\text{recon}}$ encompasses the losses in the label-encoder-decoder branch (orange branch) for the reconstruction. $\lambda_1,\lambda_2,\lambda_3$ control the weights of the three loss terms in each branch and are the same in each branch. 
$\beta$ governs the information bottleneck in $\beta$-VAE. Both branches share the decoder and the Multivariate Probit module, which in turn helps and regularizes the learning of the latent subspace where $\textbf{z}$ is embedded. The whole model can be trained end-to-end through back-propagation with Adam \cite{kingma2014adam} (see Alg. \ref{alg:algorithm}). 

\subsubsection{Interpretability of  \texorpdfstring{$\Sigma_g$}{sigmag}}

One thing special in MPVAE compared to other multi-label prediction methods is the global parameter $\Sigma_g$ (i.e. $\Sigma_r+I$). Suppose each target/label can be represented by a vector $\textbf{v}_i,i\in[1,L]$, the correlations can be captured by the inner product between $\textbf{v}_i$ and $\textbf{v}_j$. If the covariance matrix $\Sigma_g$ indeed contains such correlation, by Cholesky decomposition $\Sigma_g=VV^T$, each row of $V$ could be regarded as $\textbf{v}_i$. By some dimension reduction tricks like t-SNE, the vectors of similar targets should be close. {For  illustration purposes, we  show in the experiments section, using a real-world dataset (\textit{eBird}), that $\Sigma_g$ does capture such information.}

\section{Experiments}

\begin{small}
\begin{table}
\begin{minipage}[t]{.485\textwidth}
\scalebox{0.8}{
\begin{tabular}{@{}c@{\hspace{0.45em}}c@{\hspace{0.45em}}c@{\hspace{0.45em}}c@{\hspace{0.45em}}c@{\hspace{0.45em}}c@{\hspace{0.45em}}c@{\hspace{0.45em}}|c@{}}
\hline
Dataset & MLKNN & MLARAM & SLEEC & C2AE & DMVP & LaMP & MPVAE\\
\hline
\textit{eBird} & 0.5103 & 0.5101 & 0.2578 & 0.5007 & 0.5291 & 0.4768 & \textbf{0.5511}\\
\textit{fish} & 0.7641 & 0.5072 & 0.7790 & 0.7654 & 0.7684 & 0.7844 & \textbf{0.7881}\\
\textit{mirflickr} & 0.3826 & 0.4316 & 0.4163 & 0.5011 & 0.5105 & 0.4918 & \textbf{0.5138}\\
\textit{nuswide} & 0.3420 & 0.3964 & 0.4312 & 0.4354 & 0.4657 & 0.3760 & \textbf{0.4684}\\
\textit{yeast} & 0.6176 & 0.6292 & 0.6426 & 0.6142 & 0.6335 & 0.6242 & \textbf{0.6479}\\
\textit{scene} & 0.6913 & 0.7166 & 0.7184 & 0.6978 & 0.6886 & 0.7279 & \textbf{0.7505}\\
\textit{sider} & 0.7382 & 0.7222 & 0.5807 & 0.7682 & 0.7658 & 0.7662 & \textbf{0.7687}\\
\textit{bibtex}   & 0.1826 & 0.3530 & 0.4490 & 0.3346 & 0.4456 & 0.4469 & \textbf{0.4534}\\
\textit{delicious} & 0.2590 & 0.2670 & 0.3081 & 0.3257 & 0.3639 & 0.3720 & \textbf{0.3732}\\
\hline
\end{tabular}
}
\end{minipage}
~\\

\begin{minipage}[t]{.485\textwidth}
\scalebox{0.8}{
\begin{tabular}{@{}c@{\hspace{0.45em}}c@{\hspace{0.45em}}c@{\hspace{0.45em}}c@{\hspace{0.45em}}c@{\hspace{0.45em}}c@{\hspace{0.45em}}c@{\hspace{0.45em}}|c@{}}
\hline
Dataset & MLKNN & MLARAM & SLEEC & C2AE & DMVP & LaMP & MPVAE\\
\hline
\textit{eBird} & 0.5573 & 0.5732 & 0.4124 & 0.5459 & 0.5699 & 0.5170 & \textbf{0.5933}\\
\textit{fish} & 0.7349 & 0.5177 & 0.7563 & 0.7387 & 0.7426 & 0.7598 & \textbf{0.7648}\\
\textit{mirflickr} & 0.4149 & 0.4471 & 0.4127 & 0.5448 & 0.5499 & 0.5352 & \textbf{0.5516}\\
\textit{nuswide} & 0.3679 & 0.4151 & 0.4277 & 0.4724 & 0.4912 & 0.4720 & \textbf{0.4923}\\
\textit{yeast} & 0.6252 & 0.6350 & 0.6531 & 0.6258 & 0.6326 & 0.6407 & \textbf{0.6554}\\
\textit{scene} & 0.6667 & 0.6927 & 0.6993 & 0.7131 & 0.6935 & 0.7156 & \textbf{0.7422}\\
\textit{sider} & 0.7718 & 0.7535 & 0.6965 & 0.7978 & 0.7961 & 0.7977 & \textbf{0.8002}\\
\textit{bibtex}  & 0.1782  & 0.3645           & 0.4074           & 0.3884           & \textbf{0.4801}  & 0.4733           & 0.4800\\
\textit{delicious} & 0.2639 & 0.2734 & 0.3333 & 0.3479 & 0.3791 & 0.3868 & \textbf{0.3934}\\
\hline
\end{tabular}
}
\end{minipage}
~\\

\begin{minipage}[t]{.485\textwidth}
\scalebox{0.8}{
\begin{tabular}{@{}c@{\hspace{0.45em}}c@{\hspace{0.45em}}c@{\hspace{0.45em}}c@{\hspace{0.45em}}c@{\hspace{0.45em}}c@{\hspace{0.45em}}c@{\hspace{0.45em}}|c@{}}
\hline
Dataset & MLKNN & MLARAM & SLEEC & C2AE & DMVP & LaMP & MPVAE\\
\hline
\textit{eBird} & 0.3379 & 0.4735 & 0.3625 & 0.4260 & 0.4391 & 0.3806 & \textbf{0.4936}\\
\textit{fish} & 0.6377 & 0.4272 & 0.6570 & 0.6466 & 0.6379 & 0.6865 & \textbf{0.6925}\\
\textit{mirflickr} & 0.2660 & 0.2838 & 0.3636 & 0.3931 & 0.4193 & 0.3871 & \textbf{0.4217}\\
\textit{nuswide} & 0.0863 & 0.1565 & 0.1354 & 0.1742 & 0.1633 & 0.2031 & \textbf{0.2105}\\
\textit{yeast} & 0.4716 & 0.4484 & 0.4251 & 0.4272 & 0.4747 & 0.4802 & \textbf{0.4817}\\
\textit{scene} & 0.6932 & 0.7131 & 0.6990 & 0.7284 & 0.7160 & 0.7449 & \textbf{0.7504}\\
\textit{sider} & 0.6674 & 0.6491 & 0.5917 & 0.6674 & 0.6033 & 0.6684 & \textbf{0.6904}\\
\textit{bibtex}  & 0.0727  & 0.2267           & 0.2937           & 0.2680           & 0.3732           & 0.3763           & \textbf{0.3863}\\
\textit{delicious} & 0.0526 & 0.0739 & 0.1418 & 0.1019 & 0.1806 & \textbf{0.1951} & 0.1814\\
\hline
\end{tabular}
}
\end{minipage}
\caption{Top: example-F1 scores, Middle: micro-F1 scores, and Bottom: macro-F1 scores for different methods on all the datasets. The best scores are in bold. Each score is the average after 3 runs.}
\label{tab:f1s}
\end{table}
\end{small}
\begin{small}
\begin{table}
\scalebox{0.8}{
\begin{tabular}{@{}c@{\hspace{0.45em}}c@{\hspace{0.45em}}c@{\hspace{0.45em}}c@{\hspace{0.45em}}c@{\hspace{0.45em}}c@{\hspace{0.45em}}c@{\hspace{0.45em}}|c@{}}
\hline
Dataset            & MLKNN  & MLARAM & SLEEC  & C2AE   & DMVP   & LaMP            & MPVAE           \\
\hline
\textit{eBird}     & 0.8273 & 0.8186 & 0.8156 & 0.7712 & 0.7900 & 0.8113          & \textbf{0.8286} \\
\textit{fish}      & 0.8829 & 0.6710 & 0.8905 & 0.8840 & 0.8901 & 0.8880          & \textbf{0.8906} \\
\textit{mirflickr} & 0.8767 & 0.6337 & 0.8698 & 0.8973 & 0.8651 & 0.8969          & \textbf{0.8978} \\
\textit{nuswide}   & 0.9714 & 0.9711 & 0.9710 & 0.9725 & 0.9717 & 0.9801          & \textbf{0.9804} \\
\textit{yeast}     & 0.7835 & 0.7439 & 0.7824 & 0.7635 & 0.7808 & 0.7857          & \textbf{0.7920} \\
\textit{scene}     & 0.8633 & 0.9021 & 0.8937 & 0.8934 & 0.8748 & 0.9025          & \textbf{0.9094} \\
\textit{sider}     & 0.7146 & 0.6501 & 0.6750 & 0.7487 & 0.7387 & 0.7510          & \textbf{0.7547} \\
\textit{bibtex}    & 0.9853 & 0.9861 & 0.9818 & 0.9867 & 0.9874 & \textbf{0.9876} & 0.9875          \\
\textit{delicious} & 0.9807 & 0.9811 & 0.9815 & 0.9814 & 0.9821 & 0.9822          & \textbf{0.9824} \\
\hline
\end{tabular}
}
\caption{Hamming accuracies of different methods across all the datasets. Every accuracy is the average after 3 runs.}
\label{tab:ha}
\end{table}
\end{small}

\subsection{Datasets}
MPVAE is validated on 9 real-world datasets from a variety of fields including ecology, biology, images, texts, etc. The datasets are \textit{eBird} \cite{munson2011ebird}, \textit{North American fish} \cite{morley2018proj}, \textit{mirflickr} \cite{huiskes08}, \textit{NUS-WIDE}\footnote{We only use the 128-d cVALD features as the input.} \cite{chua2009nus}, \textit{yeast} \cite{nakai1992knowledge}, \textit{scene} \cite{boutell2004learning}, \textit{sider} \cite{kuhn2016sider}, \textit{bibtex} \cite{katakis2008multilabel}, and \textit{delicious} \cite{tsoumakas2008effective}. \textit{eBird} is a crowd-sourced bird presence-absence dataset collected from birders' observations. \textit{North American fish} (\textit{fish}) is a fish distribution dataset collected from the trawlers in the North Atlantic. \textit{yeast} is a biology database of the protein localization sites and \textit{sider} is another database of drug side-effects. \textit{mirflickr}, \textit{NUS-WIDE} (\textit{nuswide}), and \textit{scene} datasets are from the image domain. Finally, the \textit{bibtex} dataset contains a large number of BibTex files online and the \textit{delicious} dataset contains web bookmarks.

The datasets represent a large variety of different scales. The dimensionality varies from 15 (\textit{eBird}) to 1836 (\textit{bibtex}). The number of labels ranges from 6 (\textit{scene}) to 983 (\textit{delicious}). The size of the datasets could be as high as 200,000 (\textit{nuswide}), or as low as 1427 (\textit{sider}). Most datasets are available on a public website\footnote{http://mulan.sourceforge.net/datasets-mlc.html}. The rest can be found in the related papers.
If a dataset has been split \textit{a priori}, we follow those divisions. Otherwise, we separate the dataset into training (80\%), validation (10\%) and testing (10\%). The datasets are also preprocessed to fit the requirements of the input formats for the different methods. For example, we expand the input features with word embeddings for LaMP.

\subsection{Implementation and Model Comparison}

The encoders and decoder of MPVAE are parameterized by 3-layer fully connected neural networks with latent dimensionalities 512 and 256. The compared models share the same neural network structure for fair comparison, in cases where neural networks are used. The activation function in the neural networks is set to ReLU. $\Sigma_r$ is a shared learnable parameter of size $L\times L$. By default, we set $\beta=1.1, \lambda_1=\lambda_3=0.5, \lambda_2=10.0$. These default hyperparameter values are inherited from the existing well-trained DMVP \cite{chen2018end} and $\beta$-VAE \cite{higgins2017beta} models. We achieve the best performance for our own model in the neighborhood of this default set of parameters via grid search. We also use grid search to find the best learning rate, learning rate decay ratio and dropout ratio hyperparameters. Note that since not all datasets have sparse labels, $\lambda_3$ could be set to 0 or close to 0 in such scenarios. In practice, we found larger $\lambda_2$ gives better performance because the average ranking loss of multiple samples could provide good guidance for training the model. $\beta$ is the tradeoff between the capacity of the information bottleneck and the learnability of the decoder. In our experiments, the best values for $\beta$ are in the vicinity of 1.1.

\begin{figure}
\centering
\begin{minipage}{.9\linewidth}
  \centering
  \includegraphics[width=\linewidth]{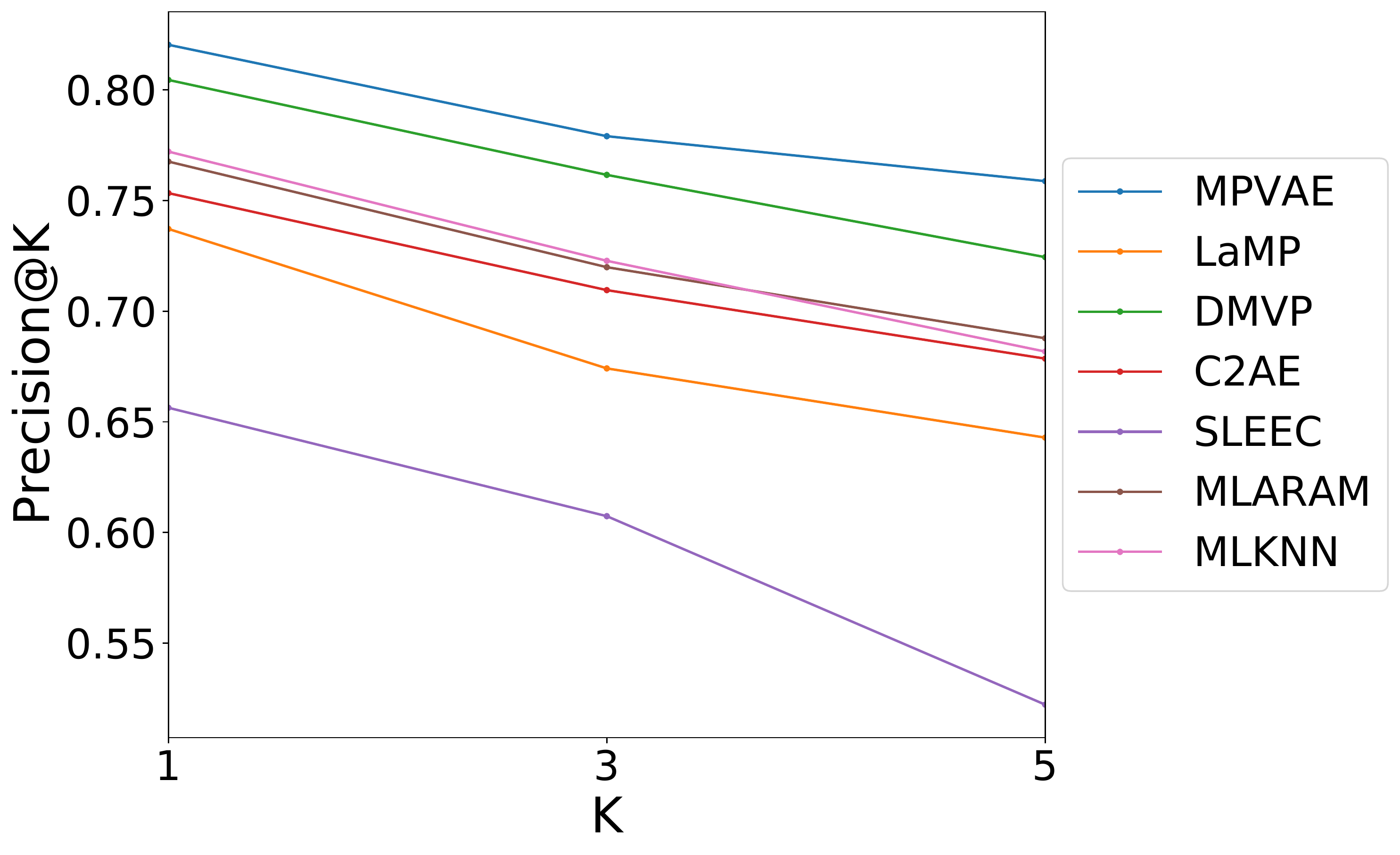}
\end{minipage}
~\\
\begin{minipage}{.9\linewidth}
  \centering
  \includegraphics[width=\linewidth]{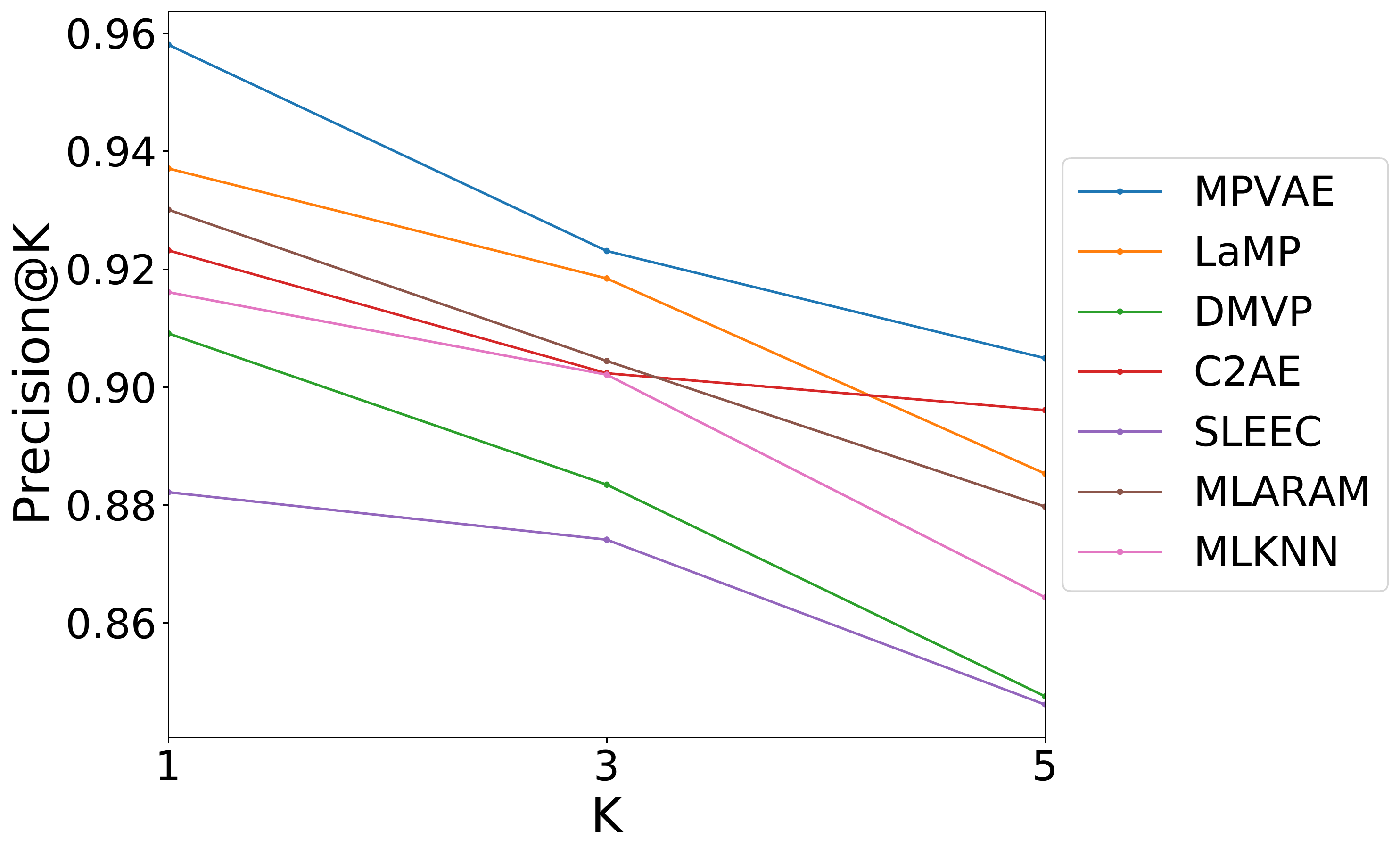}
\end{minipage}
\caption{Top: Precision@$K$ evaluations on \textit{eBird}. Bottom: Precision@$K$ evaluations on \textit{sider}.}
\label{fig:pk}
\end{figure}

MPVAE is compared with 6 other state-of-the-art methods for multi-label prediction. MLKNN \cite{zhang2007ml} is a nearest neighbor based algorithm. Bayesian inference is applied for testing. MLARAM \cite{benites2015haram} is a scalable extension to the adaptive resonance associative map neural network designed for large-scale multi-label classification. SLEEC \cite{bhatia2015sparse} learns a small ensemble of embeddings preserving local distances. It makes low-rank assumptions and can be improved with neural networks (implemented for comparison). C2AE \cite{yeh2017learning} is a recently proposed approach that learns a deep latent space through an autoencoder structure. Features and labels are encoded through deep neural networks into a latent space, where the latent embeddings for features and labels are associated by deep canonical correlation analysis (DCCA). DMVP \cite{chen2018end} is proposed for joint likelihood modeling but can be used for prediction if the trained model follows the sampling process in section 3.3 in the test phase. LaMP \cite{lanchantin2019neural} is the state-of-the-art GNN-based model for multi-label prediction. It encodes the correlations among labels to a GNN, and predicts unseen instances with the trained GNN.

\begin{figure}
\centering
\begin{minipage}{\linewidth}
  \centering
  \includegraphics[width=0.9\linewidth]{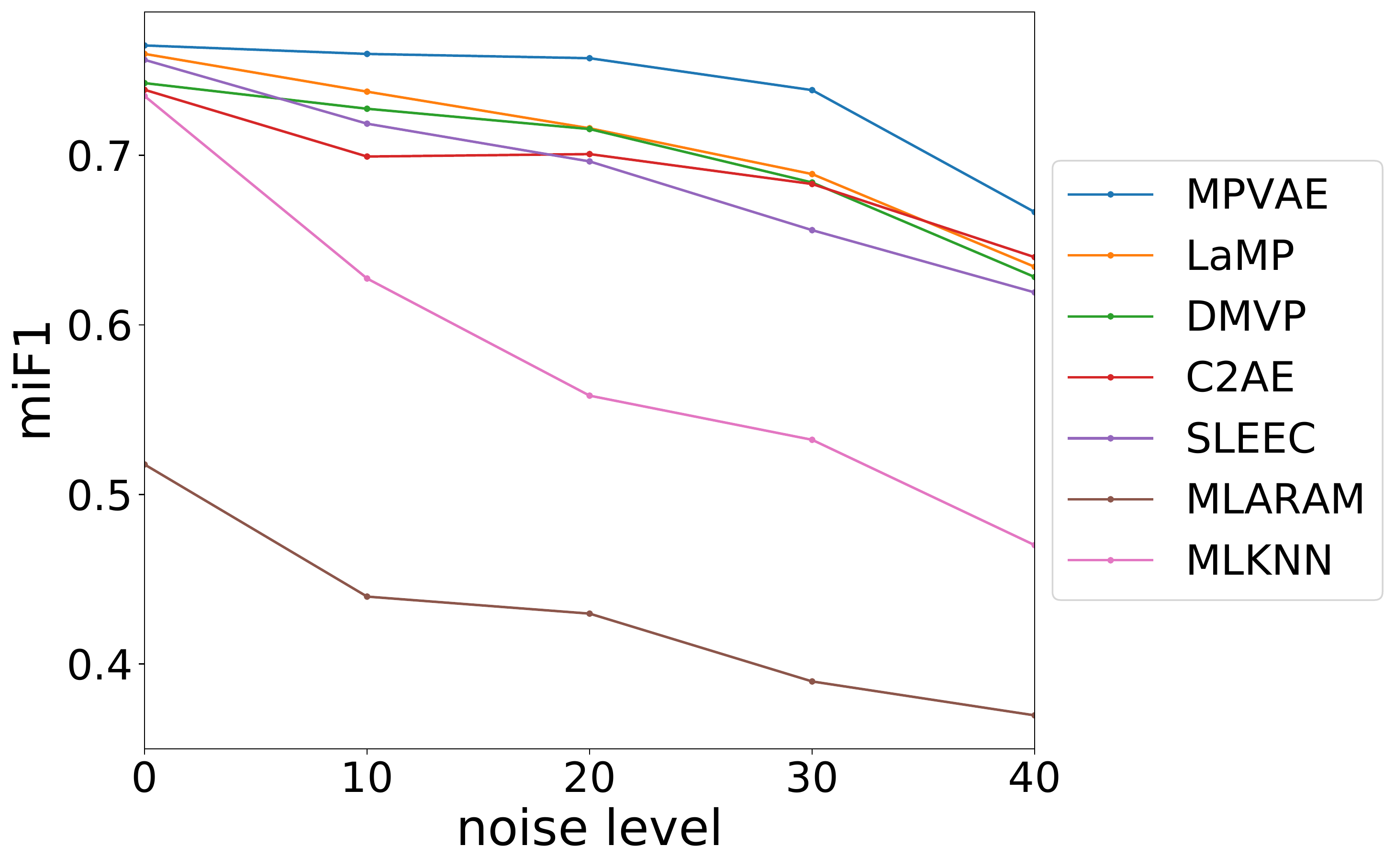}
\end{minipage}
~\\
\begin{minipage}{\linewidth}
  \centering
  \includegraphics[width=0.9\linewidth]{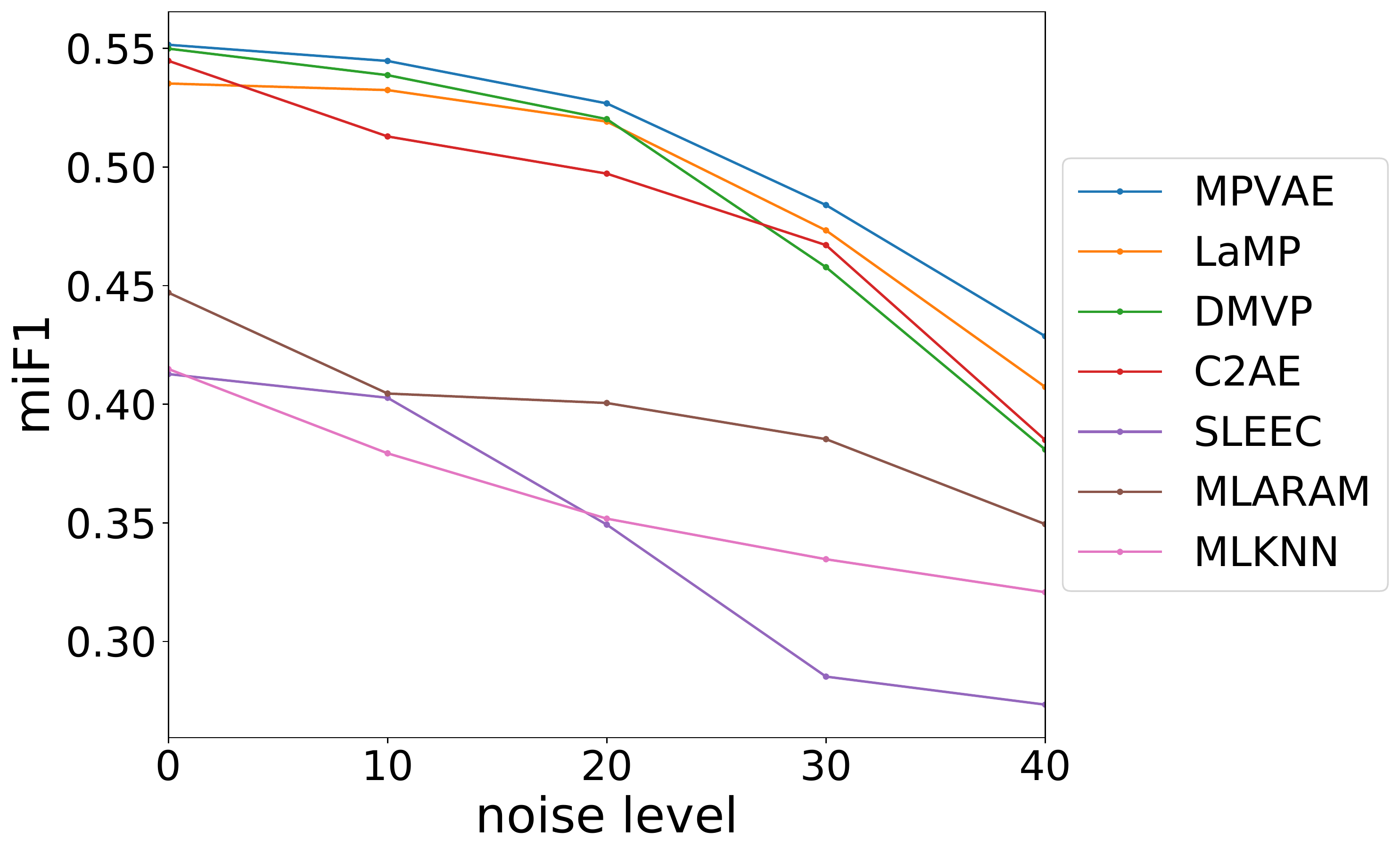}
\end{minipage}
\caption{Micro-F1 scores of the methods under different noise levels. Top: \textit{fish} dataset. Bottom: \textit{mirflickr} dataset.}
\label{fig:noisy}
\end{figure}

The major evaluation metrics for multi-label predictions are example-based F1 (example-F1), micro-averaged F1 (micro-F1) and macro-averaged F1 (macro-F1) scores.

Example-F1 measures the proportion of true positive predictions among the aggregation of the positive ground-truth labels and positive predicted labels: $\frac{1}{N_t}\sum_{i=1}^{N_t}\frac{\sum_{j=1}^L 2y_j^i\hat{y}_j^i}{\sum_{j=1}^L y_j^i+\sum_{j=1}^L\hat{y}_j^i}$, where $N_t$ is the number of test samples, $y_j^i$ is the $j$-th actual label of test sample $i$ and $\hat{y}_j^i$ is the $j$-th predicted label of test sample $i$. The \textbf{top} table in Table \ref{tab:f1s} shows the performances of different methods on all the datasets. Each F1 score is the average of 3 runs (same for the numbers in other tables and figures). MPVAE outperforms other methods on this metric. On average, MPVAE yields a 6\% improvement compared to LaMP and a 9\% improvement compared to C2AE. Micro-F1 computes the average F1 scores over all samples: $\frac{\sum_{j=1}^L\sum_{i=1}^{N_t} 2y_j^i\hat{y}_j^i}{\sum_{j=1}^L\sum_{i=1}^{N_t}[2y_j^i\hat{y}_j^i+(1-y_j^i)\hat{y}_j^i+y_j^i(1-\hat{y}_j^i)]}$. The results are given in the \textbf{middle} table in Table \ref{tab:f1s}. MPVAE is only slightly worse than DMVP on \textit{bibtex}, but compared to DMVP, MPVAE performs better by 2.5\% on average. The third F1-related metric is the macro-F1 score, which is the averaged F1 score over all labels: $\frac{1}{L}\sum_{i=1}^{L}\frac{ \sum_{i=1}^{N_t}2y_j^i\hat{y}_j^i}{\sum_{i=1}^{N_t} [2y_j^i\hat{y}_j^i+(1-y_j^i)\hat{y}_j^i+y_j^i(1-\hat{y}_j^i)]}$. Results on the \textbf{bottom} table in Table \ref{tab:f1s} illustrate that MPVAE outperforms other methods except on \textit{delicious} (second best).

\begin{figure}
\centering
\includegraphics[width=\linewidth]{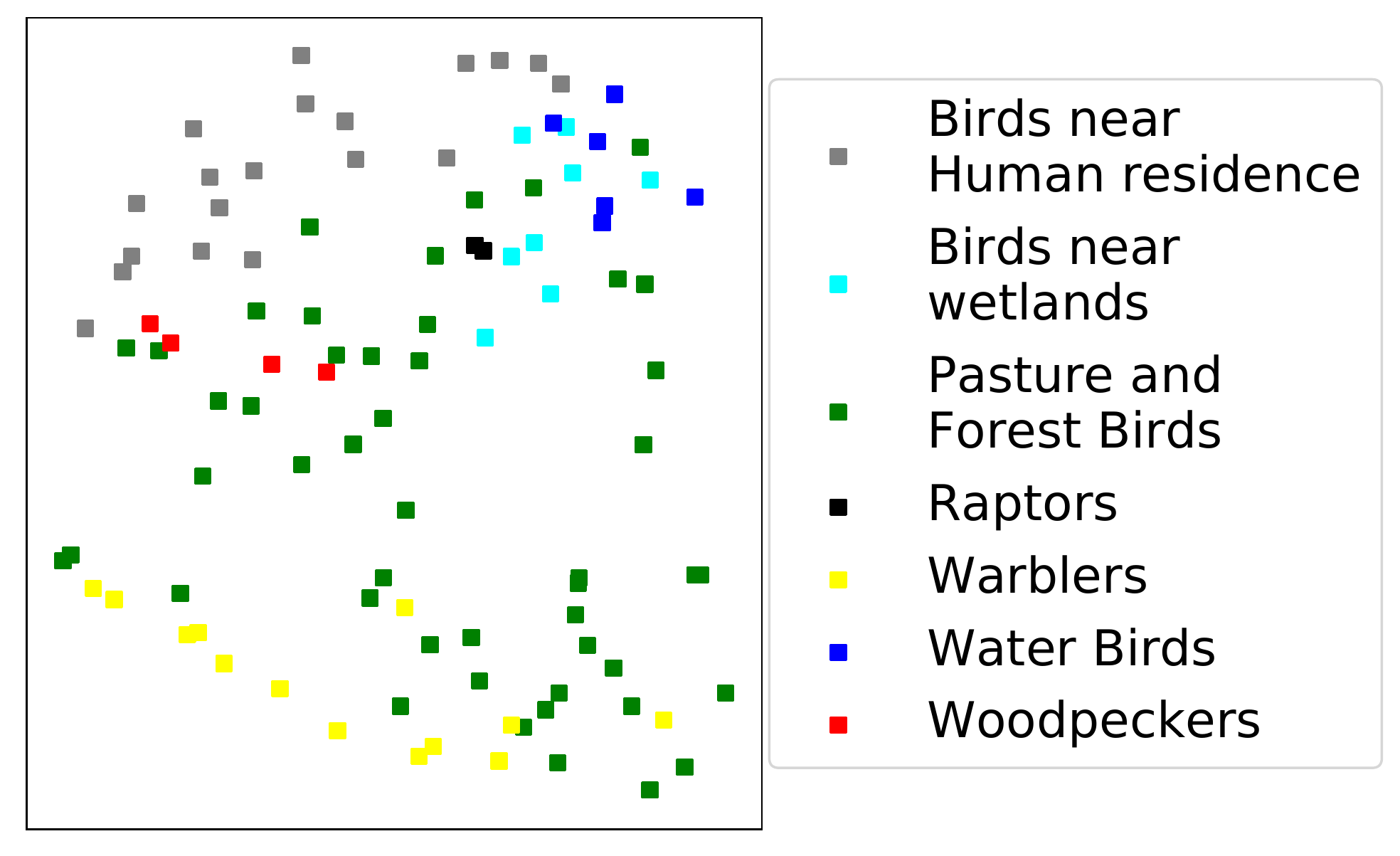}
\caption{t-SNE plot of the decomposed vectors representing each species. Different colors denote different categories of birds.}
\label{fig:cov}
\end{figure}

Besides the 3 most commonly used metrics, we test MPVAE on 2 other metrics: Hamming accuracy, and Precision@$K$. 
Hamming accuracy measures how many labels are predicted correctly: $\frac{1}{N_t}\frac{1}{L}\sum_{i=1}^{N_t}\sum_{j=1}^L\mathbb{1}[y_j^i=\hat{y}_j^i]$, regardless of pos/neg. Accuracies are collected in Table \ref{tab:ha}. The other metric Precision@$K$ is defined as the percentage of correctly predicted labels in the top-$K$ predictions. MPVAE is validated on two datasets \textit{eBird} and \textit{sider}, w.r.t. Precision@$K$ (Fig. \ref{fig:pk}). The definition, implementation and threshold selection on the validation set for the evaluation metrics follow the paper \cite{lanchantin2019neural}. 

\subsection{Noisy Labels}

Noisy labels are quite common in real-world datasets. For example, in trawl survey data of fish, the raw collected presence or absence of species might be misrecorded \cite{carton2018soda3}. Though the datasets we use have been cleaned and calibrated, the noisy setting can be reproduced by randomly flipping the labels in the training data. We tested all the methods on \textit{fish} and \textit{mirflickr} w.r.t. 4 noise levels: 10\%, 20\%, 30\% and 40\%. The comparisons are demonstrated in Fig.  \ref{fig:noisy}. As the noise level increases, MPVAE is still the most robust one. This is because when the latent dimensionality of VAE is relatively small, the model is forced to focus on the strong patterns and ignore the noise. The global covariance matrix also helps with the robustness. But as the noise level reaches 30\% and beyond, all the methods perform much worse since the noise affects the whole distribution.

\subsection{Interpreting the Covariance \texorpdfstring{$\Sigma_g$}{sigmag}}

We validate the interpretability of $\Sigma_g$ on \textit{eBird} dataset. As we mentioned in section 3.3, $\Sigma_g$ can be decomposed as $VV^T$. We regard each row of $V$ as $\textbf{v}_i$ and plot these vectors using t-SNE (see Fig. \ref{fig:cov}). One can observe that birds in the same category are clustered together. Similar clusters are also close to each other; e.g., water birds and birds near wetlands have similar embeddings. Since forest and pasture birds are the most commonly seen birds, it's not surprising that they spread across the plot. In contrast, the embeddings of rare raptors are close together. The bird categories and habits are collected from experts and professional websites\footnote{https://ebird.org/home}.

\section{Conclusion}

In this paper, we propose a disentangled Variational Autoencoder based framework incorporating covariance-aware Multivariate Probit model (MPVAE) for multi-label prediction. MPVAE comprises a feature encoder, a label encoder, a shared decoder and a Multivariate Probit model. Encoders are learned for the features and labels respectively to map them to a probabilistic subspace. The samples from the subspaces are decoded under the Multivariate Probit model to give the prediction. The disentangled $\beta$-VAE module improves the label embedding learning as well as feature embedding learning. The Multivariate Probit module provides a simple and convenient way to capture the label correlations. More importantly, we claim that the learned covariance matrix in the MP model is interpretable as shown in a real-wolrd dataset. MPVAE performs favorably against other state-of-the-art methods on 9 public datasets and remains effective under noisy settings, which verifies the usefulness and robustness of our proposed model. 

\section*{Acknowledgments}

This work is supported by National Science Foundation awards OIA-1936950 and CCF-1522054. We also want to thank the Cornell Lab of Ornithology and Gulf of Maine Research Institute for providing data, resources and advice.

\bibliographystyle{named}
\bibliography{ijcai20}

\end{document}


\maketitle

\section{Derivation Details of DMVP}

Suppose $r\sim \mathcal N(-\mu, \Sigma), z\sim \mathcal N(0, I), w\sim\mathcal N(\mu, \Sigma-I)$. Then $r=z-w$.

\begin{equation}
\begin{aligned}
    &\Phi(0|-\mu, \Sigma)\\
    =&P(r\le 0)\\
    =&P(z-w\le 0)\\
    =&P(z\le w)\\
    =&\int_{-\infty}^{+\infty}\int_{-\infty}^w p(z)p(w)dz dw\\
    =&\int_{-\infty}^{+\infty}p(w)[\int_{-\infty}^wp(z)dz]dw\\
    =&\mathbb E_{w\sim \mathcal N(\mu,\Sigma-I)}[P(z\le w|w)]\\
    =&\mathbb E_{w\sim \mathcal N(\mu,\Sigma-I)}[\prod_i P(z_i\le w_i|w_i)]\\
    =&\mathbb E_{w\sim \mathcal N(\mu,\Sigma-I)}[\prod_i \Phi(w_i|0,1)]\\
\end{aligned}
\end{equation}

$p(\cdot), P(\cdot)$ denote pdf and cdf respectively. $\Phi$ represents the cdf of a normal distribution parameterized by the mean and (co)variance.